\documentclass[11pt,letterpaper]{article}
\usepackage{cogsys}
\usepackage[T1]{fontenc}
\usepackage{times}
\usepackage[pdftex]{graphicx} 
\usepackage{listings}
\usepackage{hyperref}

\usepackage{natbib}
\cogsysheading{X}{20XX}{1-6}{X/20XX}{X/20XX}

\ShortHeadings{Open World Physics}
              {Z.\ Zeng and E.\ Davis}

\newcommand{\ifff}{\mbox{$\Leftrightarrow$}}
\newcommand{\fa}{\mbox{$\forall$}}
\newcommand{\te}{\mbox{$\exists$}}
\newcommand{\implies}{\mbox{$\Rightarrow$}}

\begin{document} 

\title{Physical Reasoning in an Open World}
 
\author{Zhuoran Zeng}{zz2017@nyu.edu}
\author{Ernest Davis} {davise@cs.nyu.edu}
\address{Dept. of Computer Science, New York University,
         New York, NY  10012, USA}
\vskip 0.2in
 
\begin{abstract}
Most work on physical reasoning, both in artificial intelligence and in 
cognitive science, has focused on closed-world reasoning, in which it is
assumed that the problem specification specifies all relevant objects and 
substance, all their relations in an initial situation, and all exogenous
events. However, in many situations, it is important to do open-world reasoning;
that is, making valid conclusions from very incomplete information. We have
implemented in Prolog an open-world reasoner for a toy microworld of
containers that can be loaded, unloaded, sealed, unsealed, carried, and dumped. 
\end{abstract}

\section{Introduction: Open vs. closed world reasoning}
Consider the following scenarios:
\begin{itemize}
\item[A.1]
You pack clothes and a tube of toothpaste in a duffel bag, you lock the 
zipper, you check it 
onto a flight to Chicago. The duffel bag is lost. Three days later, it 
turns up at the Dallas airport. It's pretty scuffed up, 
but intact. Who knows what they did to it or how it got there. 
However, if it's still locked, you can be sure 
that the clothes are still inside, but it will not be very surprising to find that there is now 
toothpaste over everything in the bag. 

\item[A.2]
In an archaeological site, a male human skeleton is found with a broken
ulna that healed. You can infer that during his lifetime, the person broke
his arm and then it healed, though you know essentially nothing else about
the person's life or what has happened to the skeleton between the time he died
and the time it was found.

\item[A.3]
A hurricane is predicted for where you live, so you take appropriate
actions: you board up the windows, bring in the lawn furniture, stock up on
water and so on. These are all sensible precautions, even though you
cannot predict exactly what will happen; e.g. which, if any, of your windows 
would break if you did not board them up.

\item[A.4]
A condominium suddenly collapses. You fear that anyone who was inside and did
not manage to escape is gravely injured or dead. Note that you can make this
inference even if you have never before heard of a condominium spontaneously
collapsing and would not have supposed that it was possible. 
\end{itemize}

From the standpoint of reasoning, what these have in common is that they
all deal with {\em open-world\/} scenarios, in which one must carry out
reasoning with very limited knowledge of the events that occurred and the 
objects that are involved.

As the above examples illustrate, open-world physical reasoning is common 
and important in real life situations. However, it has rarely been studied 
in AI literature and is almost completely unstudied in cognitive psychology.
Instead, the focus of earlier research in physical reasoning has been almost
entirely on closed-world reasoning, in which it is assumed that all the physical
entities that are relevant are enumerated, all aspects (to some fixed
level of detail) of an initial state are stated, and that all the relevant 
events are either enumerated or can be inferred from a dynamic theory.

We have developed a proof-of-concept logic-based reasoner, implemented in 
Prolog, for doing open-world reasoning about a toy world in which an
agent can manipulate containers. In our microworld, objects can be loaded 
into containers or unloaded from them; objects can be carried from one
location to another; an open container can be closed with a lid, and a lidded
container can be opened. Our system doesn't deal with shapes and sizes; any
object or set of objects can be put inside any open container.

Our program can handle examples such as the following:
\begin{itemize}
\item[B.1]  Over the interval $[T1,T2]$ object $OB$ is put into
open container $OC$. Over the interval $[T2,T3]$ container $OC$ is closed
with lid $OL$. Between $T3$ and $T4$, the lid is not opened. Infer that $OB$
is still inside $OC$ at time $T4$.
\item[B.2]  Over the interval $[T1,T2]$ object $OB$ is put into 
open container $OC$. Suppose between $T2$ and $T3$, 
$OB$ is not unloaded from $OC$ and no object is overturned (``dumped'').
Then we can safely infer that 
at $T3$, $OB$ will still remain in $OC$. 
\item[B.3]  Over the interval $[T1,T2]$ object $OB$ is put into 
open container $OC$. Suppose between $T2$ and $T3$. $OC$ is carried from location A to location B. This will not 
change the fact that, at $T3$, $OB$ is still contained in $OC$ and has been carried together with $OC$ to location B.
\end{itemize}

Equally important are the inferences that {\em cannot\/} be made in an open
world. For instance suppose that you are given that at time $T1$, object $OB$
is inside open container $OC$, and that time $T2 > T1$. In a closed world, one
can infer that $OB$ remains inside $OC$ at time $T2$, 
since there is no reason to suppose
that it has come out. In an open world, this inference cannot be made, since
$OB$ may have been removed from $OC$ in between.

The microworld and the examples are 
certainly very limited, but the analysis reveals aspects of dealing with 
open-world physical reasoning that are likely to be important in 
broader settings.

Section~\ref{secRelatedWork} will review related work. 
Section~\ref{secMicroworld} will discuss the microworld that our system
works in.
Section~\ref{secReasoning} will discuss reasoning in the abstract.
Section~\ref{secImplementation} will discuss the implementation.
Section~\ref{secFutureWork} will discuss future work.

\section{Related Work}
\label{secRelatedWork}

\subsection{Open-world physical reasoning}
Physical reasoning tasks generally involve the following elements:
\begin{itemize}
\item A number of {\em objects} or {\em substances} (broadly speaking).
These have properties and relations, some numeric, some non-numeric, some
fixed over time, some time-dependent.
\item A {\em starting situation:} the state of the world at $t=0$.
\item A {\em system trajectory:} What happens over a time interval.
\item (Optional) Some number of {\em exogenous events or actions}: Events
that take place due to external influences such as agents, that cannot
be predicted from the theory.
\item A {\em dynamic theory:} The physical laws that govern the spontaneous
evolution of the system and its response to exogenous events.
\end{itemize}

The distinction between closed and open world physical reasoning is not a binary 
dichotomy. Rather, the information available in different classes of 
reasoning problems can vary along a number of dimensions. 

\begin{itemize}
\item The numerical and geometric
information about the starting situation may be exact, approximate,
probabilistic, qualitative, or partial. (These are not
mutually exclusive.)
\item The other properties and relations that hold in the initial
situation may be completely specified or partially unspecified.
\item The objects that enter into the trajectory may be enumerated. 
\item Exogenous events may be completely specified or partially specified.
\item The dynamic theory that governs the phenomena may be exact, approximate,
probabilistic, qualitative, or partial.
\end{itemize}

The ``most closed'' form of closed world physical reasoning is 
{\em deterministic prediction}:
the starting situation
is completely specified in all aspects (up to some level of precision); 
exogenous events or boundary conditions
are likewise completely specified; the dynamic theory is complete and exact.
The reasoning task is, given the initial situation, exogenous events, and
dynamical theory, to predict the exact trajectory (again, up to some level 
of precision). ``Physics engines'', such as those that power video games, are
almost all in this category, as is the majority of scientific simulation.

In {\em probabilistic (Monte Carlo) simulation}, 
probability distributions are specified for the initial situation and the
dynamic theory is probabilistic; the task is to generate samples from the
corresponding probability distribution over the trajectories. 
In cognitive psychology, this is often called the ``noisy Newton'' approach
(Gerstenberg et al., 2012). In a {\em fully observable} scenario, the reasoner
is given a complete description of the current state; in a {\em partially
observable\/} scenario, they are given only partial or probabilistic information.
In {\em 
qualitative reasoning\/}, the initial situation and dynamic theory are given
in qualitative terms; the task, typically, is to compute the possible
trajectories in qualitative terms. In {\em inverse\/} problems, the trajectory
is given and the task is to infer the values of various numerical or geometric
parameters. One can further usefully distinguish between inverse problems 
in which the given information determines all aspects of the scenario and
those where it does not.
In {\em theory induction\/} a collection of trajectories are given, and the
task is to infer the underlying dynamical theory, or the most likely one, 
within some space of possible candidates under consideration.

\begin{table}
{\bf Knowledge-based AI physical reasoning:} \\
\hspace*{2em} {\bf Deterministic prediction:} (Funt, 1980), 
(Zickler \& Veloso, 1999). \\ 
\hspace*{2em} {\bf Probabilistic prediction (Monte Carlo):} Rare. \\
\hspace*{2em} {\bf Partially observable states:} Rare. \\
\hspace*{2em} {\bf Qualitative envisionment:}
(Forbus, 1985), (de Kleer and Brown, 1985). \\
\hspace*{2em} {\bf Inverse reasoning:}
Rare. \\
\hspace*{2em} {\bf Radically incomplete reasoning:}
(Hayes, 1979) (Davis, Marcus, and Frazier-Logue 2017). \\

{\bf Cognitive psychology:} \\
\hspace*{2em} {\bf Deterministic prediction:} 
(Schwartz \& Black, 1996),
(Smith, Battaglia, \& Vul, 2018). \\
\hspace*{2em} {\bf Probabilistic prediction (Monte Carlo):}
(Battaglia, Hamrick, \& Tenenbaum, 2013), \\
\hspace*{6em} (Gerstenberg, Goodman, Lagnado, \& Tenenbaum, 2012). \\ 
\hspace*{2em} {\bf Partially observable states:} (Sanborn, 2014). \\
\hspace*{2em} {\bf Qualitative envisionment:} 
(Forbus \& Gentner, 1997).\\
\hspace*{2em} {\bf Inverse reasoning:}
(Sanborn, Mansinghka, \& Griffiths, 2013). \\
\hspace*{6em} (Bramley, Gerstenberg, Tenenbaum,  \& Gureckis, 2018) \\ 
\hspace*{2em} {\bf Radically incomplete reasoning:} Rare. \\

{\bf Scientific computing} \\
\hspace*{2em} {\bf Deterministic prediction:} 
(Dahlquist \& Bj\"{o}rck, 2008). \\
\hspace*{2em} {\bf Probabilistic prediction (Monte Carlo):}
(Liu, 2001) \\
\hspace*{2em} {\bf Partially observable states:} ?? We don't know. \\
\hspace*{2em} {\bf Qualitative envisionment:} None.\\
\hspace*{2em} {\bf Inverse reasoning:} (Vogel, 2002) \\
\hspace*{2em} {\bf Radically incomplete reasoning:} None.
\caption{Some examples of different kinds of physical reasoning}
\label{tabReasoningTypes}
\end{table}

There is also  
large body of work on AI planning in {\em partiallly observable worlds} such as 
POMDPS in which
the starting state and subsequent states are only partially characterized
including high-level planning (e.g. Eiter et al. 2000), 
robotic planning (e.g. Hahnheide et al. 2017), and adversary games (e.g.
Bethe, 2020;
Brown and Sandholm, 2019). There is a small body of work of this kind
in the cognitive psychology literature on physical reasoning; e.g. problems
involving reasoning about collisions between objects of unknown mass (Sanborn,
2014). However, there is little work in this category in AI physical reasoning.
How much work on scientific computation with partially observable states, we
do not know.

The ``most open'' form of physical reasoning is reasoning that is {\em radically
incomplete\/} (Davis, Marcus, \& Frazier-Logue, 2017). That is, the information
given is so weak that the full trajectory {\em cannot\/} be inferred; only
some facts can be inferred. For example, in example A.1, you cannot know how
exactly the duffel bag got to Dallas or who did what to it {\em en route.}
In example B.1, you cannot know
where $OC$, $OL$, and $OB$ are located at time $T4$, 
since the lidded container may have
been moved to some other location between times $T3$ and $T4$.
Analyses based 
on monotonic logics are often appropriate to
open-world reasoning --- in monotonic logics, closed world assumptions must
be asserted explicitly --- but, previously, these have hardly been implemented.
(Situation calculus was designed primarily for closed-world planning; hence,
general open world physical reasoning tends to be awkward, even in monotonic
formulations.) 

Table~\ref{tabReasoningTypes} gives some typical references for examples of
the various types of
inference in the AI physical reasoning literature, cognitive psychology, and
scientific computing. The literature in many of these categories is large; in 
some, it is enormous.

\subsection{Reasoning about containers}
AI studies of reasoning about containers have mostly been in theoretical
studies of logical reasoning, with no implementated reasoner (Hayes, 1979;
Hayes, 1985; Davis, 2008; Davis, 2011; Davis, Marcus, \& Frazier-Logue, 2017).

In cognitive psychology, reasoning about containers has primarily been studied
in small children (Aguiar \& Baillargeon, 1998; Hespos \& Baillargeon, 2006).
It has been demonstrated (Hespos \& Baillargeon, 2001) 
that children as young as 2-1/2 months 
understand some of the physical properties of containers --- for instance, the
fact that an object cannot be put into a closed container. 

\section{The Microworld} 
\label{secMicroworld}

The microworld includes five sorts of entities: 
objects, locations, actions, times, and states (Boolean fluents).

There are five kinds of objects:
\begin{itemize}
\item {\em Closed containers}.
A closed container has an internal cavity
that may contain other objects. It cannot be opened. Objects are either 
outside or inside forever. 
\item {\em Open containers}.
An open container has an cavity with an opening on top. Other objects can be loaded
into an open container.
\item {\em Lids}.   A lid can be used to close an open container.
\item {\em Lidded containers}.  This is a pair of
an open container and a lid; it functions as a single object.
It acts as a temporary
closed container; nothing can come in or out until the lid is taken off.
\item {\em Blocks}. 
None of the above. Blocks can be put into containers but
cannot contain anything.
\end{itemize}

The idea that any kind of container can be used for any kind of content is, 
of course, a huge idealization. In reality, the question of whether A can 
serve as a container for B involves complex interactions of
spatial and physical properties.  A string shopping bag can be used as a 
container for a potato but not for a pea.  A bird cage can be used to 
contain a canary. It cannot be used to contain an ant, because the ant can 
crawl out; it cannot be used to contain a gorilla, because the gorilla is 
too large. A lucite box is a closed container as regards air, but light passes
freely in an out. The spatial and physical reasoning involved in the cases 
where all the objects involved are rigid solid objects is in principle 
characterized in theories like (Davis, Marcus, \& Frazier-Logue, 2017), though
it has not been implemented in a reasoner. Reasoning about flexible containers
such as string bags is {\em terra incognita\/}. (Physics engines can certainly
deal with simple cases of containment with rigid solid objects, though not
elegantly. However, once the containment relation becomes complex, they have
trouble; for instance, though doors are common in video games, very few game 
engines use a physically realistic model (Farokhmanesh, 2021).) For the purpose
of this initial investigation, we ignore all these issues.

The Prolog predicate
{\tt components(OC,OL,OW)} means that the lidded container $OW$ consists in the
open container $OC$ and lid $OL$.
(Throughout this paper, we will use typewriter font for Prolog and for
logical formulas. Following a common Prolog convention, variables will 
start with an upper case character; predicates, functions, and constants
will start with a lower case character.)

An object is {\em top-level\/} at a given time
if it is not inside any other object and not a 
component of a lidded container.

Locations in this theory are general areas; any number of objects may be at
a particular location. Locations have no properties, except the objects that
are at them.

\begin{figure}
\begin{center}
\includegraphics[width=6.0in]{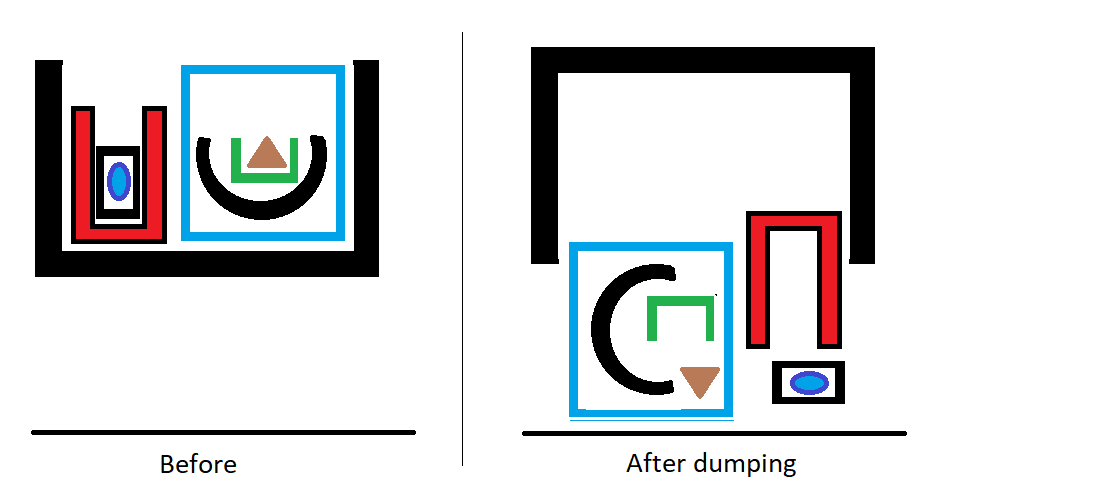}
\end{center}
\caption{Dumping a container}
\label{figDump}
\end{figure}

There are six kinds of actions:
\begin{itemize}
\item Carrying object $O$ to location $L$; in our Prolog notation,
{\tt carry(O,L)}. Any top-level object can be carried to any location.
Any objects contained inside $OA$ remains inside $OA$.

\item Loading object $O$ into open container $OC$: 
{\tt load(O,OC)}. 
$O$ and $OC$ must both
be top-level and at the same location. 
If $O$ itself is a container and $OB$
is contained in $O$, then $OB$ remains in $O$. (Containers are loaded carefully,
so that their contents do not spill).

\item Unloading object $O$ from open container $OC$: {\tt unload(O,OC)}. 
$OC$ must be top-level
and $O$ must be directly contained in $OC$. The result of unloading is that
$O$ is now top-level at the same location as $OC$.
If $O$ itself is a container and $OB$
is contained in $O$, then $OB$ remains in $O$.

\item Sealing open container $OC$ with lid $OL$ forming lidded container $OW$:
{\tt seal(OC,OL,OW)}.
$OL$ must fit on $OC$, and 
both must be top-level at the same location. As a result, the composite
container $OW$ becomes ``effective'' and $OC$ and $OL$ become ``ineffective''.

\item Unsealing lidded container $OW$ and splitting it into open container $OC$
and lid $OL$: \\
{\tt unseal(OW,OL,OC)}. $OW$ must be top-level and effective.
$OW$ becomes ineffective and $OC$ and $OL$ become effective. 

\item Dumping a container: {\tt dump(O)}. 
Intuitively, turning it upside down, and letting
the contents spill out. If container
$O$ get dumped, then, recursively, all the containers inside $O$ get dumped.
If container $OA$ gets dumped, and object $OB$ is contained in $OA$, directly
or indirectly, then $OB$ gets dumped. If $OB$ is inside $OC$ (possibly equal to
$OA$) which is a closed container or an lidded container, then $OB$ remains
inside $OC$. If $OB$ is not inside any closed container, then it falls to the 
ground outside. Figure~\ref{figDump} shows an example.
\end{itemize}
 
Five kinds of states (Boolean fluents) are used to characterize these actions: 
\begin{itemize}
    \item {\tt outsideAt(O,L)}: Object $O$ is at location $L$, not inside any
container. 
    \item {\tt directContained(OA,OC)}: object $OA$ is directly contained in object $OC$. 
\item {\tt contained(OA,OB)}. Object $OA$ is directly or indirectly contained
in $OB$. The non-reflexive transitive closure of {\tt directContained(OA,OB)}.
\item {\tt effective(O)}: Object $O$ exists and can be individually manipulated.
\item {\tt ineffective(O)}: Object $O$ either does not exist (a potential
lidded container that has not been assembled) or cannot be separately manipulated
(an open container or a lid that is part of an assembled lidded container).
\end{itemize}

Named instants of time are linearly ordered by the relation {\tt earlier(T1,T2).}
This relation means that, {\em among the instants named in the problem 
specification}, $T1$ immediately precedes $T2$; it does not mean that they
are sequential in the {\em actual} time line. In particular, any number of
unnamed events may have occured between $T1$ and $T2$. Indeed,
the theory of time is agnostic as to the kind of linear ordering of the actual
time line; the time line may 
correspond to integers, to real numbers, or are some other kind of
linear ordering. Thus, the theory could easily be extended to reasoning in a
dense or continuous model of the time line.  We do make the assumption that
there are never two events occurring simultaneously, since the causal theory
of simultaneous events tends to be much more complicated.

Following (McDermott, 1982),
we use the predicate {\tt holds(T,S)} to mean that state $S$ holds at time $T$,
and the predicate {\tt occurs(T1,T2,E)} to mean that event/action $E$ occurs
between time points $T1$ and $T2$. 

\section{Open world reasoning and the frame inference}
\label{secReasoning}
The key feature of the kind of open world reasoning that we are studying in this
project is the ability to do partial prediction over intervals in which it
is not known what events occurred. Of course, if there is no information at all
about what happened between time $TA$ and time $TB$, then the only states that
can be excluded in $TB$ are those that are entirely unreachable from the
situation in $TA$ by any sequence of events. In our microworld, the only
simple states of this kind are the contents of a closed container, which cannot
be changed. (There are also, additionally, unattainable conjunctions; 
it is impossible that
$OA$ is contained in $OB$ and $OB$ is contained in $OA$; it is impossible
that object $O$ is both at location $LA$ and location $LB$.)  

In general, to infer that a given state $S$ that holds in $TA$ will continue
to hold in $TB > TA$, it suffices to know that none of the events that
would cancel $S$ occur between $TA$ and $TB$. This can be carried out in
a logic-based system using frame axioms stated as {\em explanation closure\/}
axioms (Schubert, 1994) which specify necessary conditions for a fluent
to change values. For example, in our microworld, there is an axiom
stating that if object $OA$ is contained in a closed container
with lid $OW$ at time $T1$ and is not contained in $OW$ at time $T2$, then, at
some time in between $T1$ and $T2$, $OW$ must have been opened.

Let us first define the predicate {\tt occursWithin(TA,TB,E)} to mean that
there is an occurrence of {\tt E} that overlaps the interval {\tt [TA,TB]}.
\begin{quote}
{\tt $\fa_{\tt TA,TB,E}$ occursWithin(TA,TB,E) $\ifff$ \\
\hspace*{2em} earlier(TA,TB) $\wedge$ \\
\hspace*{2em} $\te_{\tt TC,TD}$ earlier(TC,TD) $\wedge$ earlier(TC,TB)
$\wedge$ earlier(TA,TD) $\wedge$ \\
\hspace*{4em} occurs(TC,TD,E).}
\end{quote}

We can now formally state the above frame axiom:

\begin{quote}
Frame axiom F.1: \\
{\tt $\fa_{\tt OA,OC,OL,OW,T1,T2}$ earlier(T1,T2) $\wedge$  \\
\hspace*{2em}
containerWithLid(OW) $\wedge$ components(OC,OL,OW) $\wedge$ \\
\hspace*{2em}
holds(T1,effective(OW)) $\wedge$ \\
\hspace*{2em} holds(T1,contained(OA,OW)) $\wedge$  \\
\hspace*{2em} $\neg$holds(T2,contained(OA,OW)) $\implies$ \\
\hspace*{1em}
occursWithin(T1,T2,unseal(OC,OL,OW)).} 
\end{quote}

\pagebreak

The contrapositive of F.1 is the frame inference; If $OA$ is contained in 
sealed container $OW$ at $T1$ and no unsealing of $OW$ occurs within $[T1,T2]$ 
then $OA$  is still contained in $OW$ at $T2$.

\begin{quote}
{\tt $\fa_{\tt OA,OC,OL,OW,T1,T2}$ earlier(T1,T2) $\wedge$  \\
\hspace*{2em}
containerWithLid(OW) $\wedge$ components(OC,OL,OW) $\wedge$ \\
\hspace*{2em}
holds(T1,effective(OW)) $\wedge$ \\
\hspace*{2em} holds(T1,contained(OA,OW)) $\wedge$  \\
\hspace*{4em}
$\neg$occursWithin(T1,T2,unseal(OW,OL,OC)) $\implies$ \\
\hspace*{2em}
holds(T2,contained(OA,OB)).}
\end{quote}

\section{Implementation}
\label{secImplementation}
We have implemented an inference engine for open-world reasoning for this
microworld in 
Prolog.\footnote{\url{https://github.com/Jennifercheukyin/Physical-Reasoning-in-Open-World}} 
A problem specification consists in
\begin{itemize}
\item A sequence of time instance, completely ordered under the {\tt earlier}
relation.
\item A collection of statements of the form {\tt holds(T,Q)}. Often these are
all stated at some starting time {\tt t0}, but specifications can include 
assertions about other times as well.
\item A collection of statements of the form {\tt occurs(TA,TB,E)}. The
intervals for different actions may not overlap temporally.
\item A collection of statements that specific action do not occur within
particular intervals. As we will discuss in section~\ref{secPrologKluge}, 
because of idiosyncracies of
Prolog, these take a number of different forms.
\end{itemize}

There are two key predicates characterizing change. The first is \\
{\tt condEffect(E,GOAL,QLIST)}, which means that, if $E$ occurs
from $T1$ to $T2$ and all the states in $QLIST$ hold in $T1$ then $GOAL$
will hold in $T2$. For example, in implementing example B.1, we 
use the following:

\begin{verbatim}
% If you load OB into OC, then OB is (unconditionally) inside OC.
condEffect(load(OB,OC),contained(OB,OC),[]).

% If OA is contained in OB at time T1 and you load OB into OC 
% from T1 to T2, then OA will be contained in OC at T2.
condEffect(load(OB,OC),contained(OA,OC),[contained(OA,OB)]).

% If OA is directly contained in OC at time T1 and you seal OC 
% with lid OL forming OW from T1 to T2, then OA is directly 
% contained in OW at T2
condEffect(seal(OC,OL,OW),contained(OA,OW),[contained(OA,OC)]).
\end{verbatim}

Note that the conditions are not the {\em preconditions} of action $E$, 
which must be true for
the action to be executed. In fact, our implementation of {\tt infer} does not
deal with action preconditions at all. The occurence of the action is given as a 
boundary constraint (presumably it is observed or stated), so necessarily 
the preconditions are satisfied. 
 
The second key predicate is {\tt persists(TA,TB,Q)}. This means
that, if $Q$ is true
at time $TA$ then it will still be true at time $TB$. This holds if either
(a) the problem specification asserts that {\tt occurs(TA,TB,E)} and $E$ 
does not change $Q$; or (b) if the problem specification asserts that
none of the actions that change $Q$ occur within $[TA,TB]$.

For instance, for the example below, we will use the rule, ``If 
the lidded container $OW$ is not unsealed in any interval overlapping 
$T1$ and $T2$, then
the condition `$OA$ is contained in $OW$' persists from $T1$ to $T2$.''

\begin{verbatim}
    persists(T1,T2,contained(OA,OW)) :-
           containerWithLid(OW),
           components(OC,OL,OW),
           infer(holds(T1,effective(OW))),
           notOccurs(T1,T2,unseal(OW,OL,OC)).
\end{verbatim}

(We use the Prolog predicate {\tt notOccurs(T1,T2,E)} to mean the logical
formula \\
{\tt $\neg$OccursWithin(T1,T2,E)}. The need for this is discussed in
section~\ref{secPrologKluge}.)

The reasoning engine is the query {\tt ?- infer(holds(T,Q)).} It uses 
recursive backward chaining through time. 
\begin{quote}
State $Q$ holds at time $T$ if either \\
a. The problem specification asserts {\tt holds(T,Q)}. \\
b. The problem specification asserts {\tt occurs(TX,T,E)}, the world model
asserts {\tt condEffect(E,Q,L)}, and the inference engine can recursively
verify {\tt infer(TX,G)} for every goal {\tt G} in {\tt L}. \\
c.  Let $TA$ be the time instant 
preceding $T$ in the {\tt earlier} relation. The recursive subgoal 
{\tt infer(holds(TA,Q))} succeeds and the inference engine can validate
{\tt persists(TA,T,Q).}
\end{quote}

Thus, in Prolog we have the top-level code
\begin{verbatim}
     infer(holds(T,Q)) :- holds(T,Q). % Rule 1

     infer(holds(T,Q)) :-             % Rule 2
          occurs(TA,TEND,ACT), 
          condEffect(ACT,GOAL,QLIST),
          forall(member(Q,QLIST),infer(holds(TA,Q))).

     infer(holds(T,Q)) :-             % Rule 3  
          earlier(TA,T), infer(holds(TA,Q)), persists(TA,T,Q). 
\end{verbatim}

In view of the complexity of the effects of actions like dumping in this
microworld --- both recursive and context-dependent --- the effects of actions
are individually coded, rather than being simple add lists and delete lists. 
        
\subsection{Example B.1 in Prolog}
Example B.1 discussed above can be encoded with the following 
specification:
\begin{verbatim}
block(oa). openContainer(oc). lid(ol). containerWithLid(ow).
components(oc,ol,ow).  location(la).

earlier(t0,t1). earlier(t1,t2). earlier(t2,t3).

holds(t0,outsideAt(oa,la)). 
holds(t0,outsideAt(oc,la)). holds(t0,outsideAt(ol,la)).
holds(t0,effective(oc)). holds(t0,effective(ol)). 
holds(t0,ineffective(ow)).

occurs(t0,t1,load(oa,oc)).
occurs(t1,t2,seal(oc,ol,ow)).

notOccurs(t0,t3,unsealToAnything(ow))
\end{verbatim}

(The strange form {\tt notOccurs(t0,t3,unsealToAnything(ow))} will be explained
in section~\ref{secPrologKluge}.)

The query {\tt ?- infer(holds(t3,contained(oa,ow))} succeeds, using the
following inference path: \\
Rule 3 for {\tt infer} creates the subgoal 
{\tt ?- infer(holds(t2, contained(oa,ow))). } \\
We now apply Rule 2: Since
{\tt contained(oa,ow)} is an effect of {\tt seal(oc,ol,ow)} given
the condition {\tt contained(oa,oc)}, we create the subgoal 
{\tt holds(t1,contained(oa,oc))}. \\
Applying Rule 2 again, we find that this
holds since {\tt contained(oa,oc)} is an unconditional effect of 
{\tt load(oa,oc)}. \\
Returning to the first application of Rule 3, we create
the subgoal \\
\hspace*{2em} {\tt ?- persists(t2,t3,contained(oa,ow))}. \\
Using the persistence
rule stated earlier, plus additional rules (not stated here)
that allow us to infer {\tt notOccurs(t2,t3,unseal(ow,ol,oc))} from \\
{\tt notOccurs(t0,t3,unsealToAnything(ow))} and the order on time points,
we satisfy the persistence subgoal. That completes the inference of the
top-level query \\
{\tt ?- infer(holds(t3,contained(oa,ow)))}.

By contrast, the query {\tt ?- infer(holds(t3,outsideAt(ow,la)))} fails because
it is consistent with our information that {\tt ow} is carried to some other,
unnamed, location, between {\tt t2} and {\tt t3}.

\subsection{Issues with Prolog}
\label{secPrologKluge}
We chose to implement the reasoner in Prolog, because of its efficiency, 
clarity, and familiarity.\footnote{It has been suggested that we would do
better to use an open-world logic-programming language. 
In earlier studies (Davis, Marcus, \& 
Frazier-Logue, 2017) we used
the SPASS first-order theorem prover (Weidenbach et al., 2008) but we found
that ineffective for complex chains of reasoning. Description logics typically
have an open-world semantics, but do not seem to be expressive enough for
our purposes. We will continue to look for other options; e.g. the system
described in (e.g. Jackson, Schulte \& Bj\o{}rner, 2013). However, for 
rapid prototyping, Prolog seemed a reasonable choice.}
However, the logic used in Prolog is a somewhat awkward fit
to the kind of open world reasoning that we want, as it does not support
true negation and it makes the closed-world assumption. 
Thus, to express the non-occurrence of events, we have to use a work-around.
We have to create a predicate {\tt notOccurs(T1,T2,E)} to mean that
event {\tt E} does not occur in any interval overlapping {\tt [T1,T2]}
Moreover, multiple
versions of event functions must be created for different kinds of
quantification of the variables.

For instance, suppose we want to assert that the state {\tt outsideAt(O,L)}
persists unless {\tt O} is loaded into some container {\tt OC}. 
One's first thought might be to write this as
\begin{quote}
{\tt persists(TA,T,outsideAt(O,L)) :- not occurs(TA,T,load(O,OC)).}
\end{quote}
but that, using Prolog's negation as failure, will make 
the closed world assumption; it will succeed as long as it is 
cannot be inferred that
{\tt O} is loaded into some {\tt OC}. We want to fail unless we {\em know\/}
that {\tt O} is not loaded into any {\tt OC}.

There is no direct way around this in Prolog. Instead, we introduce a new
predicate \\
{\tt notOccurs(T1,T2,E)}, meaning that we know that E does not occur in
the interval {\tt [T1,T2]}. But we are not yet out of the woods. The rule
\begin{quote}
{\tt persists(TA,T,outsideAt(O,L)) :- notOccurs(TA,T,load(O,OC)). }
\end{quote}
is still not right. The subgoal {\tt notOccurs(TA,T,load(O,OC))} 
will succeed as long as it holds for {\em any\/} binding of {\tt OC}.
What this rule states is that {\tt outsideAt(O,L)} persists as long
as there is {\em some\/} container {\tt OC} that {\tt O} is not loaded into.

A third attempt
\begin{quote}
{\tt persists(TA,T,outsideAt(O,L)) :- \\
         forall(openContainer(OC),notOccurs(TA,T,load(O,OC)). }
\end{quote}
still does not work. This will loop through all the open containers that
are explicitly mentioned in the problem specification; but in an open
world we want to avoid excluding the possibility that {\tt O} is loaded
into to some  unnamed container.

Instead, we must define a separate predicate ``loadIntoSomething(O)''
and state the rule as
\begin{quote}
{\tt persists(TA,T,outsideAt(O,L)) :-  \\
notOccurs(TA,T,loadIntoSomething(O)). }
\end{quote}
The corresponding constraint appears in the problem specification.

\subsection{Incompleteness of Inference}
The inference engine describe here is not, by any means, complete for the
microworld; that is, there are inferences valid in the microworld that 
the inference engine cannot carry out.

For instance, in our microworld, given the two facts 
{\tt occurs(t1,t2,carry(o,la))} and {\tt openContainer(oc)} it
is in fact a consequence that {\tt holds(t2,notContained(o,oc)}; only
top-level objects can be carried, so at time {\tt t1}, {\tt o} is not 
contained in {\tt oc}, and this state persists through a carry action. But
our inference methods do not include inferring that a precondition of an
action holds from the fact that the action occurs.

\section{Future Work}
\label{secFutureWork}
We plan to continue this project in a number of directions:

\begin{itemize}
\item Add functionalities of the reasoning systems
\begin{itemize}
\item Reasoning about sets of objects in addition to individual objects.
\item Incorporating some degree of geometric reasoning.
\item Adding a front end that can take input from images.
\end{itemize}
\item Characterize theoretically the power of algorithms and the difficulty
of problems in this open-world reasoning.
\item Run experiments to study this kind of reasoning in humans.
\end{itemize}

\section{Conclusions}
Intelligent action often requires reasoning that carried out in an ``open 
world'' in which an agent must come to conclusions based on knowledge that 
is significantly incomplete. By contrast, existing automated physical 
reasoning systems are almost all carry out 
reasoning that uses strong closed-world assumptions, in which initial states
and exogenous events or actions are completely specified, relative to some
As a step toward bridging this gap, we have developed, as a proof of concept, 
an initial version of an open-world reasoning system for
a toy world of containers that includes actions with complex, recursive
effects.  Unlike most physical reasoners, the reasoning system can make 
inferences despite incomplete specifications of both the starting state and
of the sequence of actions taken. The reasoning system has been implemented in 
Prolog.  The inference procedure is not logically
complete but it is sound, and works on our test examples of problems of moderate 
complexity. We plan to extend the system to richer physical theories and
other forms of partial specification.

\section*{Acknowledgements}
Thanks to Pat Little and anonymous reviewers 
for helpful feedback. This research is in part funded by
NSF award 2121102.

\section*{References}

\parindent -10pt\leftskip 10pt

\vspace{1pt}
\hspace*{-1.2em}
Aguiar, A., \& Baillargeon, R. (1998). 
Eight-and-a-half-month-old infants' reasoning about containment events. 
{\em Child Development}, {\bf 69}(3), 636-653.

\vspace{6pt}
Battaglia, P. W., Hamrick, J. B., \& Tenenbaum, J. B. (2013). 
Simulation as an engine of physical scene understanding. 
{\em Proceedings of the National Academy of Sciences,\/} 
{\bf 110}(45), 18327-18332.

\vspace{6pt}
Bethe, P. (2020). Advances in computer bridge: Techniques for a
partial-information, communication-based game. Ph.D. thesis,
Computer Science, New York University.

\vspace{6pt}
Bramley, N. R., Gerstenberg, T., Tenenbaum, J. B., \& Gureckis, T. M. 
(2018). Intuitive experimentation in the physical world. 
{\em Cognitive Psychology,} {\bf 105}, 9-38.

\vspace{6pt}
Brown, N. and Sandholm, T. (2019). Superhuman AI for multiplayer poker.
{\em Science,} {\bf 365}(6456):885-890.

\vspace{6pt}
Dahlquist, G. \& Bj\"{o}rck, \r{A}. (2008). 
{\em Numerical methods in scientific computing}, Volume I. 
Society for Industrial and Applied Mathematics.

\vspace{6pt}
Davis, E. (2008). Pouring liquids: A study in commonsense physical reasoning. 
{\em Artificial Intelligence,} {\bf 172}(12-13), 1540-1578.

\vspace{6pt}
Davis, E. (2011). How does a box work? A study in the qualitative 
dynamics of solid objects. {\em Artificial Intelligence,} {\bf 175}(1), 299-345.

\vspace{6pt}
Davis, E., Marcus, G., \& Frazier-Logue, N. (2017). 
Commonsense reasoning about containers using radically incomplete 
information. {\em Artificial Intelligence,} {\bf 248}, 46-84.

\vspace{6pt}
Eiter, T., Faber, W., Leone, N., Pfeifer, G., \& Polleres, A. (2000). 
Planning under incomplete knowledge. In {\em International Conference 
on Computational Logic} (807-821). Springer, Berlin, Heidelberg.

\vspace{6pt}
Farokhmanesh, M. (2021). Why game developers can't get a handle on doors.
{\em The Verge,} March 12, 2021. 

\vspace{6pt}
Finzi, A., Pirri, F., \& Reiter, R. (2000). Open world planning 
in the situation calculus. {\em AAAI-2000}.

\vspace{6pt}
Gerstenberg, T., Goodman, N., Lagnado, D., \& Tenenbaum, J. (2012). 
Noisy Newtons: Unifying process and dependency accounts of causal 
attribution. In {\em Proceedings of the annual meeting of the cognitive 
science society\/} (Vol. 34, No. 34).

\vspace{6pt}
Hanheide, M., G\"{o}belbecker, M., Horn, G. S., Pronobis, A., Sj\"{o}\"{o}, K., 
Aydemir, A., \ldots \& Wyatt, J. L. (2017). 
Robot task planning and explanation in open and uncertain worlds. 
{\em Artificial Intelligence,} 247, 119-150.

\vspace{6pt}
Hayes, P. (1979). The naive physics manifesto. In D. Michie (ed.), 
{\em Expert Systems in the Microelectronic Age.} 
Edinburgh: Edinburgh University Press

\vspace{6pt}
Hayes, P. (1985). Ontology for liquids. In J.~Hobbs and R.~Moore, {\em
Formal Theories of the Commonsense World.} Norwood: Ablex 71-107.

\vspace{6pt}
Hespos, S. J., \& Baillargeon, R. (2001). Reasoning about containment 
events in very young infants. {\em Cognition,} {\bf  78}(3), 207-245.

\vspace{6pt}
Hespos, S. J., \& Baillargeon, R. (2006). 
D\'{e}calage in infants' knowledge about occlusion and containment 
events: Converging evidence from action tasks. 
{\em Cognition}, {\bf 99}(2), B31-B41.

\vspace{6pt}
Jackson, E. K., Bjorner, N., \& Schulte, W. (2013). Open-world logic 
programs: A new foundation for formal specifications. 
Microsoft technical report MSR-TR-2013-55.

\vspace{6pt}
Jiang, Y., Walker, N., Hart, J., \& Stone, P. (2019). Open-world 
reasoning for service robots. In {\em Proceedings of the International 
Conference on Automated Planning and Scheduling\/} (Vol. 29, 725-733).

\vspace{6pt}
Liu, Jun S. (2001). {\em Monte Carlo Methods in Scientific Computing.}
Springer-Verlag. 

\vspace{6pt}
Sanborn, A.N. (2014). Testing Bayesian and heuristic predictions of mass
judgments of colliding objects. {\em Frontiers in Psychology}, {\bf 5}:938

\vspace{6pt}
Sanborn, A. N., Mansinghka, V. K., \& Griffiths, T. L. (2013). 
Reconciling intuitive physics and Newtonian mechanics for colliding objects. 
{\em Psychological Review}, {\bf 120}(2), 411-437.

\vspace{6pt}
Schubert, L. (1994). Explanation closure, action closure, and the Sandewall test
suite for reasoning about change. 
{\em J. Logic and Computation,} {\bf 4}5: 679-700.

\vspace{6pt}
Schwartz, D. L., \& Black, J. B. (1996). Analog imagery in mental 
model reasoning: Depictive models. {\em Cognitive Psychology,} 
{\bf 30}(2), 154-219

\vspace{6pt}
Smith, K. A., Battaglia, P. W., \& Vul, E. (2018). 
Different physical intuitions exist between tasks, not domains. 
{\em Computational Brain \& Behavior}, {\bf 1}(2), 101-118.

\vspace{6pt}
Vogel, Curtis (2002). {\em Computational Methods for Inverse Problems.}
Society for Industrial and Applied Mathematics.

\vspace{6pt}
Weidenbach, C. et al. (2009). SPASS Version 3.5. {\em 22nd Intl. Conf. 
on Automated Deduction (CADE), LNCS 5563,} 140-145

\vspace{6pt}
Zickler, S., \& Veloso, M. M. (2009). Efficient physics-based 
planning: sampling search via non-deterministic tactics and skills. 
{\em Proceedings, International Conference on Autonomous Agents and
Multi-Agent Systems,} 27-33. 

\end{document}